\pdfoutput=1

\documentclass[11pt]{article}

\usepackage[final]{acl}

\usepackage{times}
\usepackage{latexsym}
\usepackage{booktabs}
\usepackage{subcaption}
\usepackage{amsmath}
\usepackage{amssymb}
\usepackage{enumitem} 
\usepackage{graphicx}
\usepackage{multirow}
\usepackage{hyperref}

\usepackage[T1]{fontenc}

\usepackage[utf8]{inputenc}

\usepackage{microtype}

\usepackage{inconsolata}

\title{Mutual Reinforcement of LLM Dialogue Synthesis and Summarization Capabilities for Few-Shot Dialogue Summarization}

\author{
\begin{tabular}{c}
Yen-Ju Lu$^\dagger$\thanks{ Research conducted during internship at Apple. Corresponding authors: ylu125@jhu.edu, tingyao\_hu@apple.com} , Ting-Yao Hu$^*$, Hema Swetha Koppula, Hadi Pouransari,\\ Jen-Hao Rick Chang, Yin Xia, Xiang Kong, Qi Zhu,\\ Simon Wang, Oncel Tuzel, Raviteja Vemulapalli
\end{tabular}
\\
\begin{tabular}{c}
$^\dagger$Johns Hopkins University \quad Apple
\end{tabular}}

\begin{document}
\maketitle
\begin{abstract}

In this work, we propose Mutual Reinforcing Data Synthesis (MRDS) within LLMs to improve few-shot dialogue summarization task.
Unlike prior methods that require external knowledge, we mutually reinforce the LLM’s dialogue synthesis and summarization capabilities, allowing them to complement each other during training and enhance overall performances.
The dialogue synthesis capability is enhanced by directed preference optimization with preference scoring from summarization capability.
The summarization capability is enhanced by the additional high quality dialogue-summary paired data produced by the dialogue synthesis capability.
By leveraging the proposed MRDS mechanism, we elicit the internal knowledge of LLM in the format of synthetic data, and use it to augment the few-shot real training dataset. 
Empirical results demonstrate that our method improves dialogue summarization, achieving a 1.5\% increase in ROUGE scores and a 0.3\% improvement in BERT scores in few-shot settings. 
Furthermore, our method attains the highest average scores in human evaluations, surpassing both the pre-trained models and the baselines fine-tuned solely for summarization tasks.

\end{abstract}

\section{Introduction}

\begin{figure}[t]
    \centering
    \includegraphics[width=1.0\columnwidth]{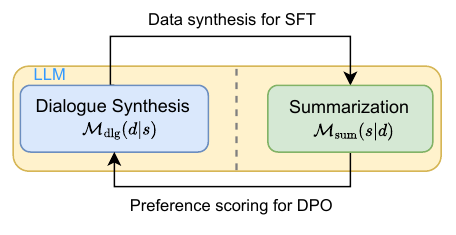}
    \caption{Mutual Reinforcement via Data Synthesis - We first leverage the summarization capability of the pretrained LLM for scoring dialogue preference pairs that are used to improve the dialogue synthesis capability with DPO, and then use the improved synthesis capability to generate SFT data for improving the summarization capability.}
    \label{fig:align}
    \vspace{-15pt}
\end{figure}

Dialogue summarization focuses on producing concise and coherent summaries of conversations in various domains such as customer service \citep{feigenblat-etal-2021-tweetsumm-dialog, zhao2021todsum}, medical consultations \citep{chintagunta-etal-2021-medically, jain2022survey, zeng-etal-2020-meddialog}, and casual interactions \citep{chen2021dialogsum,gliwa-etal-2019-samsum}. 
While state-of-the-art large language models (LLMs) such as Llama3 \citep{dubey2024llama} 
have been shown to work well for a wide variety of natural language processing tasks, their dialogue summarization performance is unsatisfactory in many target domains of interest. This could be because, these models, despite being trained on massive datasets, may have seen limited data related to dialogue summarization in specific target domains. 

Collecting a large-scale, real-world dialogue dataset and manually annotating it with concise summaries is not only time-consuming and expensive, but it also raises privacy concerns in many target domains, as conversational data is often sensitive in nature. 
Consequently, there is a growing interest in developing few-shot learning approaches that improve the dialogue summarization capabilities of pretrained LLMs using limited real dialogue-summary pairs from the target domain.


Existing works that address data scarcity for dialogue summarization assume access to either additional dialogue-summary pairs from other domains \citep{li-etal-2023-dionysus,park2024gendex,yu-etal-2021-adaptsum,zou-etal-2021-low, zhong2022dialoglm} or additional unlabeled dialogues from the target domain \citep{chen2021simple,he-etal-2024-semi} or 
teacher models that are larger and more powerful than the target models \citep{ouyang2023compositional,pham-etal-2023-select}. While these approaches have shown some performance improvements, they need access to external resources that may not be available in all scenarios. Different from these works, we focus on improving the dialogue summarization capability of a pretrained LLM using limited real dialogue-summary pairs from the target domain without relying on any additional data sources or external models.


LLMs possess a vast amount of implicit knowledge acquired during pre-training on large-scale text corpora, enabling them to generate contextually relevant text. 
Motivated by this, we introduce the framework of Mutual Reinforcement via Data Synthesis (MRDS) that harnesses the internal knowledge embedded within these models and their inherent data synthesis capability to address the data scarcity problem in dialogue summarization. Specifically, given a pretrained LLM, our method incorporates dialogue synthesis and summarization into a mutually reinforcing cycle to enhance both capabilities simultaneously (see Fig.~\ref{fig:align}). 

To improve the dialogue synthesis capability, we create a dataset of synthetic dialogue preference pairs scored by leveraging the summarization capability of the pretrained LLM, and train a LoRA \citep{hu2022lora} adapter for dialogue synthesis using direct preference optimization (DPO) with the synthetic dialogue preference pairs in addition to supervised finetuning (SFT) with the limited real dialogue-summary pairs. This ensures that the generated dialogues are coherent and closely aligned with their corresponding summaries. To improve the summarization capability, we utilize the trained dialogue synthesis adapter to generate synthetic dialogue-summary pairs, and train a LoRA adapter for summarization by performing SFT with limited real data and generated synthetic data. Effectively, the initial summarization capability of the pretrained LLM helps improve the dialogue synthesis capability which in turn helps further improve the summarization capability.

\paragraph{Major contributions:}
\begin{itemize}
  \item \textbf{Mutual reinforcement framework:} We introduce a framework that incorporates dialogue synthesis and summarization capabilities of an LLM into a mutually reinforcing cycle, and enhances both capabilities using limited target domain data.
  \item \textbf{DPO-Enhanced dialogue synthesis:} We propose to use direct preference optimization to improve the dialogue synthesis capability of a pretrained LLM, ensuring that the generated dialogues are correctly formatted, coherent, and closely aligned with their summaries.
  \item \textbf{Experimental validation:} We demonstrate consistent improvements over various alternative approaches in terms of BERT and ROUGE scores on two widely-used benchmark datasets, namely SAMSum and DialogSum. The proposed approach also achieves the best average score in human evaluations. We also present several ablation results that demonstrate the effectiveness of individual components of the proposed approach.
\end{itemize}




\section{Related Work}
\paragraph{Low-resource/Few-shot Dialogue Summarization.}
Multiple lines of methods have been proposed to address the data sparsity problem of dialogue summarization.
\citet{yu-etal-2021-adaptsum} and \citet{zou-etal-2021-low} employ datasets from multiple source domains to conduct pre-training.
\citet{he-etal-2024-semi} rely on semi-supervised learning techniques such as pseudo labeling to incorporate the additional dialogue data without summary annotation.
\citet{xie-etal-2024-shot} and \citet{zhao-etal-2022-domain} design sophisticated prompt tuning strategies, enabling cross-task knowledge transfer.
Recently, several methods have leveraged synthetic data generation using external knowledge or unlabeled datasets. For instance, GENDEX \citep{park2024gendex} generates synthetic dialogues by utilizing external knowledge bases, thereby enriching the training data and improving model performance. Similarly, compositional data augmentation proposed by \citet{ouyang2023compositional} creates new training samples by recombining existing data, enhancing diversity without additional manual annotations. Additionally, \citet{tian2024dialogue} employed a mixture-of-experts framework that integrates external knowledge to enhance summarization capabilities.
While these approaches have demonstrated performance gains, they often depend on domain-specific resources, large unlabeled datasets, or external stronger models, which may not be available or practical in few-shot settings.
This reliance on external data limits their applicability in scenarios where access to such resources is constrained. 

\paragraph{Synthetic Data from LLMs}
Many previous works have shown that LLMs are capable of synthesizing high quality training data for machine learning models. 
One line of methods primarily focus on zero-shot learning scenario \citep{ye-etal-2022-zerogen, ye-etal-2022-progen, gao2023selfguided, meng2022generating, gupta2024targen}, where they sample data from LLMs based on task related prompts, and use the synthetic data to train small, task-specific models from scratch.
Other works also demonstrate the effectiveness of synthetic data from LLM in different domains such as speech recognition \citep{corpus-synthesis}, information extraction \citep{10.1145/3477495.3531863, josifoski-etal-2023-exploiting}, text-to-SQL \citep{yang-etal-2024-synthesizing}, and dialogue state tracking \citep{kulkarni-etal-2024-synthdst, mehri-etal-2022-lad}.
Recently, several works also investigate the idea of LLM self-improvement, suggesting that the synthetic data from LLMs can improve their own instruction following abilities.
Self-instruct \citep{wang-etal-2023-self-instruct} samples from an LLM to create a synthetic prompt-response paired dataset, which can be used to finetune the original LLM.
\citet{li2024selfalignment} introduce instruction backtranslation, which obtains synthetic instruction prompts from back-translating a web scale corpus with the same LLM.
\citet{gulcehre2023reinforced}, \citet{pmlr-v235-yuan24d}, and \citet{pmlr-v235-chen24j} pay attention to the generation of responses, but utilize them in different manners.
\citet{gulcehre2023reinforced} rely on an external scoring function to obtain the reward of synthetic responses.
\citet{pmlr-v235-yuan24d} propose a self-rewarding framework, using the LLM to score the response generated from itself.
\citet{pmlr-v235-chen24j} design a self-play mechanism, finetuning the LLM to distinguish the responses generated by the itself and human responses.
%

\section{Methodology}



\begin{figure}[t]
    \centering
    \includegraphics[width=1.0\columnwidth]{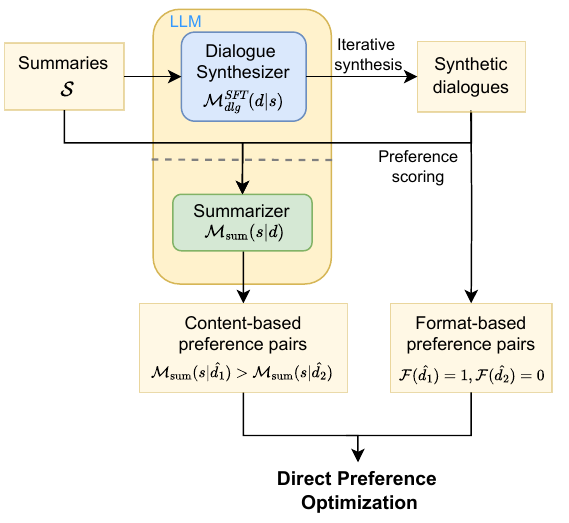}
    \caption{
    Preference pairs for DPO - Given a summary set \( \mathcal{S} \), the LLM generates synthetic dialogues which are evaluated based on content alignment and format correctness. Content-based preferences leverage the LLM's own summarization capability to assess how well the synthesized dialogues align with the input summaries. Format-based preferences ensure that the dialogues follow proper formatting.
    }
    \label{fig:dpo}
\end{figure}


\paragraph{Notations} Let $\mathcal{D}_p$ denote the limited real dataset of dialogue-summary pairs, and $\mathcal{S}$ denote the set of summaries in $\mathcal{D}_p$.

Given a pretrained LLM, we characterize its dialogue synthesis and summarization capabilities using two separate LoRA \citep{hu2022lora} adapters. 
First, we leverage the summarization capability of the pretrained LLM to score synthetic dialogue preference pairs, and use this data to train the dialogue synthesis adapter using DPO in addition to SFT on real data $\mathcal{D}_p$. Then, we leverage the trained dialogue synthesis adapter to generate synthetic dialogue-summary pairs that are used (in addition to real data $\mathcal{D}_p$) to train the summarization adapter.
Effectively, the synthesis and summarization capabilities help improve each other without using any external models or additional real data.


 
\subsection{Dialogue Synthesizer Training}
\label{sec:dialogue}
The dialogue synthesizer (pretrained LLM with dialogue synthesis adapter) takes a summary $s$ as input and generates a dialogue $\hat{d}$. 
A straightforward way to train this model is to perform SFT with the real dialogue-summary pairs $\mathcal{D}_p$.
However, when the amount of such training data is limited, the resulting dialogue synthesizer $\mathcal{M}_{dlg}^{SFT}$ often generates poor quality dialogues in terms of both dialogue format and content (see Table~\ref{tab:synthesis_sft_sample}).

To encourage the dialogue synthesizer to generate higher quality dialogues, we construct a synthetic dataset of preference pairs $\{s, \hat{d}_1, \hat{d}_2\ |\ s \in \mathcal{S}, \mathcal{P}(\hat{d}_1) >\  \mathcal{P}(\hat{d}_2)\}$, where $\mathcal{P}$ denotes a dialogue quality scoring function, and train the synthesizer using DPO. Specifically, we generate two types of preference pairs using two different quality scoring functions, one focusing on the dialogue format and the other focusing on the dialogue content (see Fig.~\ref{fig:dpo}). Both the preferred and rejected dialogues in these pairs are synthetic dialogues generated using the SFT-trained dialogue synthesizer $\mathcal{M}_{dlg}^{SFT}$. 

 
\paragraph{Format-based Preference Pairs}
Since the SFT-trained synthesizer $\mathcal{M}_{dlg}^{SFT}$ often generates dialogues with several formatting errors, we develop an iterative dialogue synthesis (IDS) method that generates correctly-formatted dialogues.
First, we generate a dialogue conditioned on the input summary and check it for format errors. We discard the portion of the dialogue after the first detected error,
concatenate the remaining correctly-formatted partial dialogue to the initial input prompt and give it as input to the dialogue synthesizer to complete the dialogue. This process is repeated until we get a dialogue without formatting errors. Table~\ref{tab:iter} shows an example run of this IDS process. 

For a given summary, we generate multiple dialogues with format errors by directly sampling from the SFT-trained synthesizer $\mathcal{M}_{dlg}^{SFT}$ and multiple clean dialogues by repeating the above IDS process several times. We use these samples to form preference pairs
$\{s, \hat{d}_1, \hat{d}_2\ |\ s \in \mathcal{S},  
 \mathcal{F}(\hat{d}_1) = 1,\   \mathcal{F}(\hat{d}_2) = 0\}$
, where $\mathcal{F}$ denotes the format check-based binary scoring function (1 for clean dialogues, 0 for those with errors).
When trained with these preference pairs, the dialogue synthesizer learns to generate dialogues in correct format. Table~\ref{tab:preference_format_example} shows an example of format-based preference pair.


\paragraph{Content-based Preference Pairs}
A dialogue $\hat{d}$ generated by a well-trained dialogue synthesizer should have high content alignment with the summary $s$ used to generate $\hat{d}$. We leverage the summarization capability of the pretrained LLM to measure this alignment. The main motivation is that if we summarize the synthetic dialogue $\hat{d}$, the corresponding summary $s$ should have high probability as the summarization output. Based on this, we use the likelihood $\mathcal{M}_{sum}(s|\hat{d})$ of the summary $s$ conditioned on the dialogue $\hat{d}$ measured by the pretrained LLM as the alignment score.

To encourage the dialogue synthesizer to generate dialogues that have high content alignment with the input summaries, we construct a dataset of preference pairs using the content alignment score $\mathcal{M}_{sum}(s|\hat{d})$. 
Specifically, for each summary $s$, we first generate multiple clean dialogues from the SFT-trained dialogue synthesizer $\mathcal{M}_{dlg}^{SFT}$ following the IDS process described above, and then pick the dialogues with the best and least content alignment scores to form preference pairs $\{s, \hat{d}_1, \hat{d}_2\ |\ s \in \mathcal{S},
\mathcal{M}_{sum}(s|\hat{d}_1) >\  \mathcal{M}_{sum}(s|\hat{d}_2)\}$. Table~\ref{tab:preference_content_example} shows an example of content-based preference pair.

Instead of using the pretrained LLM, we also explored training the summarization adapter on the limited real data $\mathcal{D}_p$ and using it for alignment scoring to generate DPO training data. However, we did not observe significant improvements in the final summarization results (after training the dialogue synthesizer with DPO, generating synthetic dialogues, and using them to train the final summarization adapter).

\paragraph{Training with DPO and SFT}
In addition to DPO with synthetic preference pairs, we also use SFT with the limited real data $\mathcal{D}_p$ to train the dialogue synthesizer. We accumulate gradients from both losses in each optimization step. While DPO with synthetic data encourages the model to generate contextually accurate dialogues in correct format, it does not explicitly encourage the model to generate dialogues that mimic the target distribution represented by real data $\mathcal{D}_p$. By combining DPO with SFT on real data, we encourage the model to generate dialogues closer to the target distribution while being contextually accurate and better-formatted.

\subsection{Summarizer Training}
\label{sec:summ_train}
The summarization model (pretrained LLM with summarization adapter) is trained using SFT with limited real data $\mathcal{D}_p$ and additional synthetic dialogue-summary pairs.

\paragraph{Synthetic Summary Generation}
To generate synthetic summaries that mimic the distribution represented by the real summary set $\mathcal{S}$, we first extract a (2-3 words) topic for each summary $s \in \mathcal{S}$ using the pretrained LLM. 
Then, for each topic, we generate multiple new synthetic summaries by using a topic-based summary synthesizer. 
This synthesizer is obtained by supervised LoRA finetuning of the pretrained LLM using the real summaries in $\mathcal{S}$ and the corresponding extracted topics. 

\paragraph{Synthetic Dialogue Generation}
For each synthetic summary, we generate a dialogue by directly sampling from the dialogue synthesizer that has been trained with both DPO and SFT, as explained in the previous section. 
Since this improved synthesizer already generates dialogues with high quality, we do not use the costly iterative synthesis approach in this step.

\paragraph{Training with Synthetic and Real Data}
Models trained only on limited real data tend to overfit quickly. While combining real and synthetic data samples in each minibatch could address this issue to some extent, finding the perfect ratio between the two data sources is challenging and highly dependent on the quality of the synthetic data. Moreover, using a fixed ratio of real and synthetic data samples throughout the training may not be optimal. If the minibatch is dominated by synthetic data, then the model may inherit the artifacts present in the synthetic data, and if the minibatch is dominated by real data, the model may start to overfit before taking full advantage of the synthetic data. 

To address these issues, we follow a two-stage training strategy, where we train only on synthetic data in the first stage and only on real data in the second stage. The synthetic-only first stage allows the model to learn general dialogue summarization skills without the risk of quickly overfitting on limited real data. The real-only second stage allows the model to adapt to the distribution of the real data mitigating the artifacts learned from synthetic data. Following this two-stage approach, we make effective use of both synthetic and real data, resulting in a more accurate summarization model.


\section{Experiments}




\paragraph{Datasets} We experiment with two widely-used benchmark datasets, namely SAMSum~\citep{gliwa-etal-2019-samsum} and DialogSum~\citep{chen2021dialogsum}. The SAMSum dataset contains over 16,000 casual conversations mimicking everyday chats among friends and family. The DialogSum dataset includes about 13,460 face-to-face spoken dialogues between friends, colleagues, and between service providers and customers, covering various daily-life topics. These datasets also provide a human-written summary for each dialogue. Since this work focuses on few-shot settings, for each dataset, we experiment with either 100 or 300 dialogue-summary pairs as the few-shot training dataset $\mathcal{D}_p$.






\subsection{Alternative Methods}



We compare our MRDS method with several dialogue summarization alternative approaches, divided into two categories: methods using the pre-trained Llama3 model without fine-tuning and methods fine-tuned on real or synthetic data.
\paragraph{Pre-trained Methods}
\begin{itemize} [label=-, topsep=0pt, itemsep=1pt, parsep=0pt, partopsep=0pt]
\item \textbf{Zero-shot:} Zero-shot summarization performance of Llama3. 
\item \textbf{ICL:} Summarization performance of Llama3 using in-context learning with $k=7$ examples.~\footnote{We experimented with different number of in-context examples and $k=7$ worked best.}
\end{itemize}
\paragraph{Fine-tuned Methods}
\begin{itemize} [label=-, topsep=0pt, itemsep=1pt, parsep=0pt, partopsep=0pt]
\item \textbf{Real only}: Fine-tuning with real data only.
\item \textbf{SFT}: Two-stage training using synthetic dialogues generated by the SFT dialogue synthesizer.
\item \textbf{SFT + Post-processing}: Two-stage training using synthetic dialogues from the SFT dialogue synthesizer, enhanced with Iterative Dialogue Synthesis (IDS) and content alignment filtering.
\end{itemize}

\subsection{Implementation Details}
We use Llama3-8B-Instruct \citep{dubey2024llama} as the pretrained base LLM. We use a rank of 16 and an alpha of 32 for all LoRA adapters, and keep the base model parameters frozen while training the LoRA adaptors. Table~\ref{tab:prompt} shows all the prompts used for topic extraction, topic-based  summary synthesis, summary-based dialogue synthesis, and dialogue summarization tasks. All presented results are averaged over three runs.

\paragraph{Dialogue Summarization}
For the baseline model trained exclusively on real data, we optimized the hyperparameters and applied the same settings to all subsequent experiments for consistency. Our training strategy includes a batch size of 10 and a maximum learning rate of $2.0 \times 10^{-4}$ with a warmup over the first 50 batches. We use the \texttt{ReduceLROnPlateau} scheduler with a patience of 5 and a reduction factor of 0.7. Training is stopped if the loss does not improve for 100 steps. We select the best checkpoint based on the validation loss obtained during the real data training phase.

In synthetic data experiments, we employ a two-stage training approach using the same hyperparameters. In the first phase, we train exclusively on synthetic data until the learning rate reduces to $2.0 \times 10^{-5}$, effectively serving as a pre-training phase. In the second phase, we apply the same training strategy as in the real-only experiments to ensure a fair comparison.

\paragraph{Dialogue Synthesis}
For the dialogue synthesizer trained with SFT only, we use a learning rate of $2.0 \times 10^{-4}$ along with the \texttt{ReduceLROnPlateau} scheduler. The batch size and other hyperparameters are the same as those used for dialogue summarization. When training the synthesizer using both SFT and DPO, we start from the SFT checkpoint. In this combined training, we use a batch size of four for DPO and one for SFT, jointly updating the dialogue synthesizer by combining the losses from both objectives. A fixed learning rate of $1 \times 10^{-5}$ is used during this phase. We validate the synthesizer checkpoints on the official validation set of the dataset, evaluating both format correctness and summarization cross-entropy loss. We select the checkpoint with the lowest summarization CE loss, ensuring at least 85\% format correctness. Detailed training hyperparameters are provided in Table~\ref{tab:hyp}.

\section{Results}

\begin{table*}[ht]
\centering
\caption{Comparison of Summarization Methods on 100 and 300 shots.}
\label{tab:summarization}
\begin{tabular}{lcccc|cccc}
\toprule
\multirow{2}{*}{Approach} & \multicolumn{4}{c}{SAMSum} & \multicolumn{4}{c}{DialogSum} \\
\cmidrule(lr){2-5} \cmidrule(lr){6-9}
& R-1 & R-2 & R-L & BERTScore & R-1 & R-2 & R-L & BERTScore \\
\midrule
Zero shot & 31.3 & 12.3 & 23.9 & 81.2 & 28.2 & 10.0 & 21.4 & 81.6\\
ICL (k=7) & 39.5 & 17.6 & 30.9 & 83.2 & 31.4 & 11.9 & 24.5 & 83.1\\
\midrule
\textbf{100 Real shots} \\
\midrule
Real only & 50.9 & 26.5 & 42.6 & 86.6 & 44.0 & 18.2 & 36.0 & 86.8 \\ 
SFT & 50.9 & 26.5 & 42.6 & 86.5 & 45.1 & 19.0 & 36.9 & 86.9 \\ 
SFT + Post-processing & 51.8 & 27.3 & \textbf{43.5} & 86.7 & 44.7 & 18.8 & 36.5 & 87.1 \\
MRDS (ours) & \textbf{52.1} & \textbf{27.5} & 43.4 & \textbf{86.8} & \textbf{45.5} & \textbf{19.3} & \textbf{37.2} & \textbf{87.2} \\
\midrule
\textbf{300 Real shots} \\
\midrule
Real only & 51.1 & 26.9 & 42.8 & 86.5 & 45.2 & 19.5 & 37.3 & 87.2 \\
SFT & 52.1 & 27.6 & 43.7 & 86.8 & 45.9 & 19.8 & 37.7 & 87.2 \\ 
SFT + Post-processing & \textbf{52.7} & 28.1 & 44.1 & \textbf{87.0} & 46.1 & 20.0 & 37.6 & 87.3 \\
MRDS (ours) & \textbf{52.7} & \textbf{28.3} & \textbf{44.4} & \textbf{87.0} & \textbf{47.0} & \textbf{20.4} & \textbf{38.6} & \textbf{87.5} \\

\bottomrule
\end{tabular}
\end{table*}

\subsection{Dialogue Summarization}
We conducted experiments on the SAMSum and DialogSum datasets to evaluate the effectiveness of our proposed mutual reinforcing data synthesis (MRDS) method. The results are presented in Table~\ref{tab:summarization}, comparing various summarization approaches under 100-shot and 300-shot settings using metrics such as ROUGE-1 (R-1), ROUGE-2 (R-2), ROUGE-L (R-L) \citep{lin-2004-rouge}, and BERTScore \citep{bert-score}.

In the zero-shot setting, the pre-trained LLM (Zero shot) achieves R-1 scores of 31.3 on SAMSum and 28.2 on DialogSum. In-context learning approach (ICL) improves the R-1 score on SAMSum to 39.5 and DialogSum to 31.4, demonstrating efficiency of ICL in low-resource scenarios.

When fine-tuning with 100 real shots, the real-only method significantly improves performance over zero-shot methods, achieving R-1 scores of 50.9 on SAMSum and 44.0 on DialogSum. 
Incorporating dialogues from the SFT model maintains similar performance, while post-processing techniques (SFT + Post-processing) further enhance results, increasing R-1 scores to 51.8 on SAMSum and 44.7 on DialogSum, indicating the effectiveness of IDS and content alignment filtering. 
In the 300-shot setting, all methods benefit from additional training data. 
The real-only method reaches R-1 scores of 51.1 on SAMSum and 45.2 on DialogSum, with {SFT} showing incremental gains, and {SFT + Post-processing} achieving R-1 scores of 52.7 on SAMSum and 46.1 on DialogSum.

Our proposed MRDS method outperforms all approaches in both 100-shot and 300-shot settings. 
In the 100-shot setting, MRDS achieves the highest R-1 scores of 52.1 on SAMSum and 45.5 on DialogSum, along with improvements in R-2, R-L, and BERTScore metrics, highlighting its ability to leverage synthesized data effectively. 
In the 300-shot setting, MRDS continues to deliver the best performance, matching the top R-1 score of 52.7 on SAMSum and setting a new high of 47.0 on DialogSum. 
The method consistently delivers superior ROUGE and BERTScore values, highlighting its robustness and scalability with increased data. This shows that MRDS not only improves efficiency by eliminating post-processing steps but also significantly boosts summarization performance.

\begin{table}[th!]
\caption{Human evaluation results on SAMSum with 300-shot data, covering informativeness (inf.), faithfulness (fai.), fluency (flu.), redundancy (red.), and the average (ave.) of the four scores.}
\label{tab:sub}
\centering
\begin{tabular}{lcccc|c}
\toprule
     & \textbf{Inf.} & \textbf{Fai.} & \textbf{Flu.} & \textbf{Red.} & \textbf{Ave.} \\
\midrule
\textbf{Zero}          & \textbf{1.95} & 1.50 & \underline{1.94} & 0.83 & 1.56 \\
\textbf{ICL}           & \underline{1.65} & 1.69 & \textbf{2.00} & 1.23 & 1.64 \\
\textbf{Real}          & 1.32 & 1.61 & \textbf{2.00} & 1.94 & 1.72 \\
\textbf{MRDS}           & 1.55 & \textbf{1.88} & \underline{1.94} & \textbf{2.00} & \textbf{1.84} \\
\bottomrule
\end{tabular}
\end{table}

\begin{table}[htp]
\centering
\caption{Comparison of efficiency for dialogue synthesis with (Dialogues/hour).}
\label{tab:eff}
\begin{tabular}{lcc}
\toprule
{Approach} & {100 shots} & {300 shots}  \\
\midrule
\multicolumn{3}{l}{\textbf{SAMSum}} \\
\midrule
Post-Processing & 63 & 37.5  \\
MRDS & \textbf{550} & \textbf{2900}  \\
\midrule
\multicolumn{3}{l}{\textbf{DialogSum}} \\
\midrule 
Post-Processing & 85 & 66.7  \\
MRDS & \textbf{315} & \textbf{2500} \\
\bottomrule
\end{tabular}
\end{table}

\subsection{Human Evaluation}
\label{sec:he}

Table~\ref{tab:sub} presents the results of human evaluations, comparing four groups of summaries: two from the pretrained instructed Llama3 model—zero-shot results (Zero) and in-context learning (ICL)—and two from the fine-tuned Llama3-based summarization model: trained on real data only (Real) and trained with the proposed MRDS approach, which combines real and synthetic summaries and dialogues. Five human evaluators assessed the summaries based on informativeness, faithfulness, fluency, and redundancy, using a scale from zero to two, following the evaluation protocol from \citep{xie-etal-2024-shot}. The average scores across the four metrics are also reported.

For the pre-trained models, while the zero-shot model (Zero) achieves the highest score in informativeness (1.95), it scores poorly in redundancy (0.83), indicating a tendency to produce overly long summaries by including too much information --undesirable in summarization tasks. 
The in-context learning model (ICL) shows slight improvements in faithfulness (1.69 vs. 1.50) and redundancy (1.23 vs. 0.83) compared to zero-shot, indicating that it generates more concise and faithful summaries but still inherits some limitations of the pre-trained model.
For the fine-tuned models, the Real approach outperforms the pre-trained models in overall average score (1.72 vs. 1.56 for Zero and 1.64 for ICL), demonstrating the benefit of fine-tuning with real data in improving summary quality.
However, our proposed MRDS method achieves the highest average score (1.84), outperforming both the real-only and pre-trained models, particularly in faithfulness and redundancy. 
This suggests that incorporating synthesized data helps the model produce more precise, concise summaries. The results highlight that our MRDS approach significantly enhances summarization quality from the perspective of human evaluators. Example summaries from different approaches are shown in Table~\ref{tab:summ_example_1}.

\subsection{Dialogue Synthesis Efficiency}
\label{sec:result_dia}


We compared the efficiency of the post-processing approach—which includes iterative synthesis and content alignment filtering—with the DPO-based MRDS approach (Table~\ref{tab:eff}). In the 100-shot scenarios, although MRDS still required iterative synthesis due to less consistent format correctness with limited data, it achieved 550 dialogues per hour on SAMSum compared to 63 dialogues per hour with post-processing—an 8.7-fold improvement. In the 300-shot scenarios, MRDS significantly increased throughput, generating 2,900 dialogues per hour on SAMSum versus 37.5 dialogues per hour with post-processing—a 77-fold improvement—by eliminating the need for IDS or content alignment filtering due to enhanced format correctness and summarization alignment.




\subsection{Model Analysis}

\begin{table*}[ht]
\centering
\caption{Comparison of different summaries with synthesis data on DialogSum 100 and 300 real shots. Synthesis dialogues are generated by SFT-trained dialogue synthesizer with IDS and content alignment filtering.}
\label{tab:summaries}
\begin{tabular}{lcccc|cccc}
\toprule
\multirow{2}{*}{Approach} & \multicolumn{4}{c}{100 Real shots} & \multicolumn{4}{c}{300 Real shots} \\
\cmidrule(lr){2-5} \cmidrule(lr){6-9}
& R-1 & R-2 & R-L & BERTScore & R-1 & R-2 & R-L & BERTScore \\
\midrule
Real only & 44.0 & 18.2 & 36.0 & 86.8 & 45.2 & 19.5 & 37.3 & 87.2 \\
\midrule
\multicolumn{5}{l}{\textbf{Fixed Ratio Training}} \\
\midrule
Same Summ & 42.8 & 17.4 & 35.3 & 86.4 & 44.6 & 19.0 & 36.7 & 86.7 \\
Synthetic Summ & 44.1 & 18.2 & 36.1 & 86.7 & 44.0 & 18.2 & 35.5 &  86.6 \\
Unseen Summ & 44.5 & 18.8 & 36.3 & 87.0 & 46.6 & {20.5} & 38.1 & {87.5} \\
\midrule
\multicolumn{5}{l}{\textbf{Two-Stage Training}} \\
\midrule
Same Summ & 42.5 & 16.8 & 34.6 & 86.3 & 45.8 & 19.9 & 37.6 & 87.2 \\
Synthetic Summ & 44.7 & 18.8 & 36.5 & 87.1 & {46.1} & {20.0} & {37.6} & {87.3}   \\
Unseen Summ &  45.3 & 19.7 & 37.4 & 87.0 &  {46.8} & {20.8} & {38.8}  & {87.6}  \\
\bottomrule
\end{tabular}
\end{table*}

\paragraph{Summaries Effectiveness.}
To evaluate the impact of different summary types on dialogue synthesis, we experimented with two training strategies: (1) a fixed 1:1 ratio of real to synthesized data and (2) a two-stage training approach, training first on synthesized data and then switching to real data. We tested these strategies using three types of summaries: the same summaries (identical to those in the real data), synthetic summaries generated from our synthesis process in Sec.~\ref{sec:summ_train}, and unseen summaries drawn from additional training data.

Table~\ref{tab:summaries} presents the results. Under the fixed ratio strategy, using the same summaries did not yield any improvement over the real-only approach, likely due to overfitting to the limited real summaries. Synthetic summaries provided some improvement in the 100-shot setting but introduced artifacts that degraded performance in the 300-shot setting. This suggests that mixing synthetic and real data in a fixed ratio can lead to the model learning undesirable patterns from the synthetic data as the amount of real data increases.

In contrast, the two-stage training approach produced better results. Training with synthetic summaries first allowed the model to learn general patterns from the synthesized data before fine-tuning on real data, which improved performance in both the 100-shot and 300-shot settings. Specifically, using synthetic summaries in two-stage training significantly outperformed using the same summaries, mitigating overfitting and enhancing generalization. This highlights the value of synthetic summaries in the more effective two-stage training framework.
Additionally, incorporating unseen summaries improved both training strategies, but the two-stage training still provided superior results. 

\paragraph{Dialogue Generation and Filtering.}

\begin{table}[th]
\centering
\caption{Ablation study for dialogue generation, filtering, and DPO on unseen summaries. The training is conducted in two stages training.}
\label{tab:filter}
\begin{tabular}{lcccc}
\toprule
\multirow{2}{*}{Approach}  & \multicolumn{4}{c}{DialogSum 300 shots}  \\
\cmidrule(lr){2-5} 
& R-1 & R-2 & R-L & B-S   \\
\midrule
Real only & 45.2 & 19.5 & 37.3 & 87.2  \\
\midrule
\multicolumn{5}{l}{\textbf{Unseen Summaries}} \\
\midrule
w/o dialogue  & 45.8 & 19.9 & 37.3 & 87.2 \\
w/o filtering & 46.3 & 20.1 & 37.9 & 87.3 \\
w/ filtering &  \underline{46.8} & \textbf{20.8} & \textbf{38.8}  & \textbf{87.6}  \\
\midrule
MRDS & \textbf{47.0} & \underline{20.7} & \underline{38.7} & \underline{87.5}  \\
\bottomrule
\end{tabular}
\end{table}

To evaluate the benefits of generating and filtering dialogues, we experimented with unseen summaries using the two-stage training strategy. In the first stage, we tested three types of synthetic data: summaries only, summaries with unfiltered dialogues, and approaches involving content alignment filtering or DPO training (MRDS), as shown in Table~\ref{tab:filter}. We found that training with summaries alone improved upon the real-only approach; however, adding dialogues to the training data further enhanced the results. Finally, the models utilizing filtering or DPO outperformed all other methods, demonstrating that including dialogues with filtering and DPO effectively improves the final outcomes.




\section{Conclusion}
We introduce a novel approach that mutually reinforces dialogue synthesis and summarization capabilities of a large language model (LLM) to improve few-shot dialogue summarization without relying on external data.
By leveraging the dialogue synthesis capability enhanced by DPO, we synthesize well-formatted, coherent dialogues to augment the few-shot real dataset.
Furthermore, the two-stage training strategy effectively incorporated synthesized dialogues without introducing artifacts, improving summarization accuracy.
Empirical results demonstrated significant improvements: a 1.5\% increase in ROUGE scores, a 0.3\% improvement in BERT scores.
Human evaluations confirmed that our method outperforms the real-only baseline and, in certain aspects, surpasses human-annotated ground truth summaries. 
Our approach offers a practical solution for real-world applications with limited data, utilizing the model's inherent capabilities without external resources. 
Future work could extend this self-alignment framework to other NLP tasks affected by data scarcity and explore its integration with larger or more diverse LLM architectures.

\section{Limitations}
Comparing to the baseline method, training summarization adapters on few shot real data only, our method requires additional computation cost for data synthesis adapter training and sampling.
Also, our method makes an assumption that part of the internal knowledge of LLM is useful for the target domain, which might be incorrect for a highly specialized target domain.

\section{Ethics Statement}

Our research introduces Mutual Reinforcing Data Synthesis (MRDS) within LLMs to enhance few-shot dialogue summarization tasks. While advancing natural language processing, we acknowledge ethical considerations associated with our methodology.
By leveraging synthetic data generated by LLMs, we reduce reliance on large-scale real-world datasets that may contain sensitive or personally identifiable information (PII). We strive to protect user privacy and adhere to data protection regulations by using publicly available datasets (CC BY-NC-ND 4.0 and CC BY-NC-SA 4.0) and implementing data anonymization techniques.

Our method utilizes LLMs pre-trained on vast corpora that might contain biases and stereotypes. Although we train our synthesis and summarization models on public datasets with daily conversations, we recognize that biases may persist. We encourage future work to identify and reduce biases in synthetic data generation and in models trained on such data.
We are committed to transparency. All experimental details—including data preprocessing, model configurations, and evaluation metrics—are thoroughly documented to ensure reproducibility and allow critical assessment by the research community. By openly sharing our methods and findings, we aim to foster collaboration and uphold ethical standards in AI development.





\bibliography{anthology,custom}

\appendix

\section{Appendix}
\label{sec:appendix}
\subsection{Prompt Templates}
Table \ref{tab:prompt} shows the prompt templates we use in this work for different purposes.

\begin{table*}[]
\begin{tabular}{l}
\hline
\textbf{Topic extraction}\\ \hline
\begin{tabular}[c]{@{}l@{}}Please determine the main topic of the provided summary.\\ The topic should be a brief phrase that captures the essence of the summary.\\ Instructions:\\ - Ensure the topic is around 2 words in length.\\ - It should clearly reflect the core idea of the summary.\\ - Avoid specific details and names; the topic should be general enough to apply to various summaries.\\ - Use precise and specific language.\\ Summary:  {[}summary{]}\\ Topic:\end{tabular}\\ \hline
\textbf{Summary synthesis given topic}\\ \hline
\begin{tabular}[c]{@{}l@{}}Please write a summary for a document based on the provided topic.\\ Instructions:\\ - The summary should be approximately {[}word count{]} words in length.\\ - Ensure the summary captures the main idea related to the topic.\\ Topic: {[}topic{]}\\ Summary:\end{tabular}\\ \hline
\textbf{Dialogue synthesis given summary}\\ \hline
\begin{tabular}[c]{@{}l@{}}Please write a dialogue based on the given summary. \\ Each line should start with the anonymous speaker\'s index followed by a colon.\\ Instruction:\\ - Use the {[}num of speaker{]} speakers in the dialogue: \\ \#1, \#2 ...\\ - Ensure each line starts with the speaker's index followed by a colon.\\ - Write the dialogue clearly and naturally.\\- The dialogue should be around {[}num of turns{]} turns and {[}num of words in dialogue{]} words in length. \\ Summary: {[}summary{]}\end{tabular} \\ \hline
\textbf{Dialogue summarization}\\ \hline
\begin{tabular}[c]{@{}l@{}}Document: {[}Dialogue{]}\\ Summarize the provided document\end{tabular}\\ \hline
\textbf{Dialogue Summarization (zero shot)}\\ \hline
\begin{tabular}[c]{@{}l@{}}Dialogue: {[}Dialogue{]}\\ Summarize the provided dialogue.\end{tabular}\\ \hline
\textbf{Dialogue Summarization (in-context learning)} \\ \hline
\begin{tabular}[c]{@{}l@{}}Your task is to summarize the provided dialogue. The following are some examples: \\ Example 1:\\ Dialogue: {[}Example dialogue 1{]}\\ Summary: {[}Example summary 1{]}\\ \\ Example 2:\\ Dialogue: {[}Example dialogue 2{]}\\ Summary: {[}Example summary 2{]}\\ …\\  Now, here is the target dialogue: \\ {[}Dialogue{]}\\ Summarize the target dialogue.\end{tabular}\\ \hline
\textbf{Dialogue Summarization (zero shot + length control)}\\ \hline
\begin{tabular}[c]{@{}l@{}}Dialogue: {[}Dialogue{]}\\ Summarize the provided dialogue. \\ The summary should be around {[}average word count of summaries in training set{]} words in length.\end{tabular} \\ \hline
\end{tabular}
\caption{Prompt templates.}
\label{tab:prompt}
\end{table*}

\subsection{Examples of Summarization Result}
Table~\ref{tab:summ_example_1} presents an example dialogue along with summaries produced by various methods, including the human-labeled ground truth summary, the output from the zero-shot pre-trained model, summaries generated using in-context learning, and those fine-tuned with real data only as well as with the MRDS approach.

\begin{table*}[]
\centering
\begin{tabular}{l}
\hline
\textbf{Input dialogue}\\ \hline
Avery: You went to Ethan's house? \\
David: yeah I had to babysit \\
Avery: Aww, how do you babysit, just curious \\
David: I had to go through a lot :/ \\
Avery: Was his sister naughty \\
David: Tooo much \\
Avery: Lol \\
David: I will just refuse net time :/ \\
Avery: As you wish \\
David: :/ \\
Avery: I just got his text \\
David: What is he saying\\
Avery: He is asking me to say thanks to you\\
David: yeah whatever<3\\
Avery: He was saying that your phone was switched off\\
David: Yeah i have just turned it on\\
Avery: I have told him about that\\
David: k\\
Avery: Gotta go now \\
\hline
\textbf{Ground Truth Summary} \\ \hline
David was looking after Ethan's sister. Ethan is grateful. David won't do it again. \\
\hline
\textbf{Generated Summary (Zero Shot)} \\ \hline
The dialogue is between Avery and David, discussing David's experience babysitting Ethan's sister. \\
David mentions that the sister was "too much" and that he will refuse to babysit again. \\
Avery shares that Ethan sent a text asking Avery to thank David, \\
and also mentions that David's phone was switched off. \\
David confirms that he just turned it on and the conversation ends with Avery saying they have to go.\\
\hline
\textbf{Generated Summary (In-context Learning)} \\ \hline
David babysat at Ethan's house and had a difficult time. \\ Avery asks how he managed it and David says it was a lot of work. \\ Avery asks if Ethan's sister was naughty and David confirms that she was. \\ David jokes that he will refuse to babysit again. \\ Avery mentions Ethan sent a text asking Avery to thank David, and David agrees to let Avery do so.\\ Avery also mentions that Ethan's phone was switched off, which David had just turned back on. \\
\hline
\textbf{Generated Summary (Real only)} \\ \hline
David had to babysit Ethan's sister. \\
\hline
\textbf{Generated Summary (MRDS, Ours)} \\ \hline
David had to babysit Ethan's sister. Avery got a text from Ethan saying thanks to David.\\
\hline
\end{tabular}
\caption{Summarization results from our method and baselines.}
\label{tab:summ_example_1}
\end{table*}

\subsection{Synthetic Dialogue from SFT Trained Synthesizer}
To show the limited capability of dialogue synthesizer trained on few shot real data, we provide an example of low quality synthetic dialogue in Table \ref{tab:synthesis_sft_sample}.

\begin{table*}[]
\centering
\begin{tabular}{l}
\hline
\textbf{Summary $s$:} \\
\#2 can't join \#1 and Mike for a game because \\
he is studying.\\
\hline
\begin{tabular}[c]{@{}l@{}}\textbf{Generated dialogue $\hat{d}$}:\\ 
\#1: So will you come at 2?\\ 
\#2: \textcolor{red}{sure bro,} what time you gonna be with Mike?\\ 
\#1: about 20 min.00, Mike will call you\\ 
\textcolor{blue}{2: he is going}\\ 
\textcolor{blue}{3 no way, he's still sleeping hehehe} \\ 
\textcolor{blue}{2} \textcolor{red}{he's studying}\\ 
\textcolor{blue}{1: and?}\\  
\textcolor{blue}{not coming}\end{tabular} \\ \hline 
\end{tabular}
\caption{Low quality dialogue generated by SFT-trained dialogue synthesizer. \#1 and \#2 are anonymized speaker names. The text highlighted in \textcolor{red}{red} shows content that is not aligned with the summary (\#2 can't join and \#2 is studying instead of Mike). The lines highlighted in \textcolor{blue}{blue} are format errors, since they do not start with a proper speaker name (\#1,  \#2, ...).}
\label{tab:synthesis_sft_sample}
\end{table*}

\subsection{Preference Pair Examples}
Tables~\ref{tab:preference_content_example} and~\ref{tab:preference_format_example} provide examples of content-based and format-based preference pairs, respectively. In Table~\ref{tab:preference_content_example}, the summarization model scores the content alignment of the dialogues using $\mathcal{M}_{sum}(s|\hat{d})$, selecting the best and worst ones as the preferred and rejected dialogues. In Table~\ref{tab:preference_format_example}, the preferred dialogues are synthesized using IDS, where $\mathcal{F}(\hat{d}) = 1$, while the rejected dialogues are generated through the SFT synthesizer's one-shot synthesis, where $\mathcal{F}(\hat{d}) = 0$.



\begin{table*}[]
\centering
\begin{tabular}{l}
\hline
\begin{tabular}[c]{@{}l@{}}\textbf{Summary:} \\ On Sunday \#1 and \#2 are meeting with Dominica and James who got married recently.\end{tabular}\\ 
\hline
\begin{tabular}[c]{@{}l@{}}\textbf{Preferred dialogue:}\\
\#1: Hahaha we must meet Dominica tomorrow with James. \\
\#2: yep \\
\#1: I dont think she will be very available after the marriage.\\
\#2: ye ye, she's very busy\\
\#1: I asked :) :) <file\_gifile>\\
\#2: so 8  o'clock?\\
\#1: sounds super
\end{tabular}\\
\hline
\begin{tabular}[c]{@{}l@{}}\textbf{Rejected dialogue:}\\
\#1: are u free next Sunday?\\
\#1 file\_gifiles : yeah\\
\#2: ok \\
\#1: wanna meet Domin?\\
\#2: ok\\
\textcolor{blue}{\# fileifile} \\
\textcolor{blue}{\# gileile} \\
\textcolor{blue}{\# : ok: <file :)}\\
\textcolor{blue}{file :) <file :)  : :file :) <file < :)}\\
\textcolor{blue}{\# Ok, ok :) <file :)}\\
\textcolor{blue}{ :) it sounds cool}
\end{tabular} \\ \hline
\end{tabular}
\caption{Example of a format-based preference pair. \#1 and \#2 are anonymized speaker names. The lines highlighted in \textcolor{blue}{blue} are format errors, since they do not start with a proper speaker name (\#1,  \#2).}
\label{tab:preference_format_example}
\end{table*}

\begin{table*}[]
\centering
\begin{tabular}{l}
\hline
\begin{tabular}[c]{@{}l@{}}\textbf{Summary:} \\ \#1 is outside waiting for \#2, who is late.\end{tabular}\\ 
\hline
\begin{tabular}[c]{@{}l@{}}\textbf{Preferred dialogue:}\\
\#1: Hey, it's \#1. I'm sitting outside where are you?.\\
\#2: \textless{}file\_giffile\textgreater\\
\#1: You're late! And I was waiting for you there.\end{tabular}\\
\hline
\begin{tabular}[c]{@{}l@{}}\textbf{Rejected dialogue:}\\
\#1: I can't come to the meeting\\
\#2: what did?\\
\#1: I have not enough money for cab. I must wait for \#1\\
\#2: \#1 is waiting for us!\\
\#1: how can I get there?\\
\#2: use bus, it's the fastest, come back me\end{tabular} \\ \hline
\end{tabular}
\caption{Example of a content-based preference pair. \#1 and \#2 are anonymized speaker names.}
\label{tab:preference_content_example}
\end{table*}

\subsection{Data Formatting and Anonymization}

To simplify the training of the data synthesizer, we perform anonymization preprocessing on both summaries and dialogues. Specifically, we extract the number of speakers and their names from the dialogues as metadata and replace the names in both dialogues and summaries with uniform identifiers (e.g., "\#1", "\#2").

\subsubsection{Data Formatting}
\label{sec:format}
To ensure that the synthetic data adheres to the correct format, we have established a set of formatting rules:

\begin{itemize}[label=-, topsep=0pt, itemsep=1pt, parsep=0pt, partopsep=0pt]
\item \textbf{Speaker Identity}: each sentence in the dialogues must begin with a speaker identifier followed by a colon (e.g., "\#1:", "\#2:").
\item \textbf{Consistency of Names}: Synthetic dialogues and summaries should not contain incorrect or extra names—for example, "\#4" in dialogue-summary pairs with fewer than four speakers, or a "\#" symbol without a number.
\item \textbf{Inclusion of Speaker Names}: Synthetic summaries must contain at least one anonymized speaker name.
\end{itemize}

For summary synthesis, we discard any summaries that do not comply with these formatting rules. For dialogue synthesis, we apply Iterative Dialogue Synthesis (IDS) to ensure that the synthetic dialogues conform to the correct format.



\subsubsection{Iterative Dialogue Synthesis}
Table~\ref{tab:iter} illustrates an example of iterative dialogue synthesis, showing partial inputs given to the synthesizer and the corresponding synthesized outputs. During the synthesis process, we check for formatting issues as described in Sec.~\ref{sec:format}. If an error is found, the flawed sentence, along with all subsequent sentences, is discarded, and the correctly formatted segment (e.g., "\#1:", "\#2:") is used to re-synthesize the dialogue.

\begin{table*}[]
\centering
\begin{tabular}{l}
\hline
\textbf{Input summary}\\ \hline
Both \#1 and \#2 forgot about Monica's birthday today. \\ As they do not know what present to buy for her,  \#2 will phone Monica's boyfriend and ask his advice.\\ 
\hline 
\textbf{First partial response}\\ \hline
\textcolor{orange}{\textbf{\#1:}}\\
\hline 
\textbf{First synthesis result}\\ \hline
\textbf{\#1:} 
\textcolor{blue}{I forgot about Monicas birthday today haha} \\
\textcolor{blue}{\#2: Me to, me too}\\
\textcolor{blue}{\#1: what now?}\\
\textcolor{blue}{\#2: we should make something?} \\
\textcolor{blue}{\#1: what do you propose to buy?}\\
\textcolor{blue}{\#1: I'm asking my brother's opinion, he knows her a bit more}\\
\textcolor{blue}{\#2: same here, I will phone her boyfried}\\
\textcolor{blue}{\#1: you must have more info}\\
\textcolor{blue}{\textit{\underline{2: ya, I will get some advice}}}\\
\textcolor{blue}{\textit{1:}}\\
\textcolor{blue}{\textit{I asked and we decided to buy her something}}\\
\textcolor{blue}{\textit{2.: 1 can not see.}}\\
\textcolor{blue}{\textit{5. This 2 1 3’s got 1 6.75.}} \\ 

\hline 
\textbf{Second partial response}\\ \hline
\#1: I forgot about Monicas birthday today haha\\
\#2: Me to, me too\\
\#1: what now?\\
\#2: we should make something? \\
\#1: what do you propose to buy?\\
\#1: I'm asking my brother's opinion, he knows her a bit more\\
\#2: same here, I will phone her boyfried\\
\#1: you must have more info\\
\textcolor{orange}{\textbf{\#2:}}\\

\hline
\textbf{Second synthesis result}\\ \hline
\#1: I forgot about Monicas birthday today haha\\
\#2: Me to, me too\\
\#1: what now?\\
\#2: we should make something? \\
\#1: what do you propose to buy?\\
\#1: I'm asking my brother's opinion, he knows her a bit more\\
\#2: same here, I will phone her boyfried\\
\#1: you must have more info\\
\textbf{\#2:} \textcolor{blue}{sure}\\
\textcolor{blue}{\#1: I am already looking for some ideas}\\
\textcolor{blue}{\#2: it looks hopeless}\\
\textcolor{blue}{\#2: I will call and he phone}\\
\textcolor{blue}{\#1: yes }\\
\hline
\end{tabular}
\caption{An example of iterative dialogue synthesis (IDS), condensed for brevity. In each step, the correctly-formatted partial response is concatenated to the initial dialogue synthesis prompt and given as input for generation. Newly generated text appears in blue, while discarded text is italicized. In each step, we also add an anonymized speaker name (shown in yellow) at the end of the partial response before using it for generation. We always use \#1 as the first partial response, and randomly choose an anonymized speaker in the subsequent steps.}
\label{tab:iter}
\end{table*}

\subsubsection{Synthetic Dialogue Recovery}
After data synthesis, we restore the anonymized dialogues and summaries before training the summarization model by replacing placeholders with real names using metadata extracted during the anonymization process. For synthetic summaries that lack original metadata, we seed the synthesis process with metadata from the real data in the few-shot training set, then replace the placeholders accordingly. Alternatively, we can generate random names and speaker numbers to populate the placeholders.


\subsection{Human Evaluation Details}
We hired five machine learning researchers, and received their consent to report the results from their annotation work.
We adopt the human evaluation method proposed in the previous work \citep{xie-etal-2024-shot}.
For completeness, we describe the 4 metrics (informativeness, faithfulness, fluency and redundancy) and the corresponding instructions in Table \ref{tab:human_eval_instruct}.

\begin{table*}[]
\centering
\begin{tabular}{l}
\hline
\textbf{Informativeness}\\ \hline
Whether the critical information in the dialogue is missed in the summary: \\
*0: lots of the critical information in the dialogue is missed; \\
*1: a small amount of the critical information in the dialogue is missed;\\
*2: no critical information in the dialogue is missed.\\
\hline
\textbf{Faithfulness}\\ \hline
Whether the information presented in the summary is factually incorrect or unmentioned \\
according to the dialogue: \\
*0: lots of the information presented in the summary is factually incorrect or unmentioned; \\
*1: a small amount of the information presented in the summary is factually incorrect or unmentioned;\\
*2: no information presented in the summary is factually incorrect or unmentioned.\\
\hline
\textbf{Fluency}\\ \hline
Whether the sentences in the summary are ungrammatical or ill-formed: \\
*0: lots of the sentences in the summary are ungrammatical or ill-formed; \\
*1: a small amount of the sentences in the summary are ungrammatical or ill-formed;\\
*2: no sentence in the summary is ungrammatical or ill-formed.\\
\hline
\textbf{Redundancy}\\ \hline
 Whether the expressions of the summary can be simplified: \\
*0: lots of the expressions of the summary can be simplified; \\
*1: a small amount of the expressions of the summary can be simplified;\\
*2: no expression of the summary can be simplified.\\
\hline
\end{tabular}
\caption{Human evaluation metrics and their corresponding instructions}
\label{tab:human_eval_instruct}
\end{table*}

\subsection{Training Hyperparameters}
Table~\ref{tab:hyp} presents the hyperparameters used for the summarization and dialogue synthesis models. We conducted an extensive search for hyperparameters on the summarization model trained exclusively with real data. All other SFT training follows the same set of hyperparameters. The LR threshold for the two-stage summarization approach refers to the minimum learning rate reached during the first stage of training with synthetic data. Once this threshold is met, the second stage begins using real data, applying the same hyperparameters as those used for the real-only summarization model.

The DPO synthesis model is initialized from the checkpoints of the SFT synthesis model and is trained using both DPO and SFT loss. The batch size for DPO training is four, randomly sampled from two preference sets: format-based and content-based preference pairs. Additionally, SFT training data with a batch size of one is included in the process, and the combined losses are used to update the synthesis model.

\begin{table*}[htbp]
\centering
\caption{Hyper-parameters used for training dialogue summarization, two-stage dialogue summarization, dialogue/summary synthesis, and DPO-based dialogue synthesis models.}
\label{tab:hyp}
\begin{tabular}{@{} lcccc @{}}
\toprule
 & Summarization & Two-stages Summ. & Synthesis & DPO Syn. \\
\midrule
Loss & CE & CE & CE & DPO and CE \\
Batch Size & 10 & 10 & 10 & {\small 4 DPO \& 1 SFT}  \\
Learning Rate & 2.0e-4 & 2.0e-4 & 2.0e-4 & 1.0e-5 \\ 
Optimizer & ReduceLROnPlateau  & ReduceLROnPlateau  & ReduceLROnPlateau & Fixed LR \\ 
Validation Steps & 2 & 2 & 2 & 20 \\
Patience & 5 & 5 & 5 & - \\
Factor & 0.7 & 0.7 & 0.7 & - \\
Warmup Steps &  50 & 50 & 50 & - \\ 
Early Stopping &50 & 50 & 50 & - \\ 
LR Threshold & - & 2.0e-5 & - & - \\\hline
LoRA Rank &  \multicolumn{4}{c}{16} \\  
LoRA Alpha & \multicolumn{4}{c}{32} \\ 
Dropout Rate & \multicolumn{4}{c}{0.4} \\
\bottomrule
\end{tabular}
\end{table*}

\subsection{Dialogue Synthesis Quality}

\begin{table*}[htp]
\centering
\begin{tabular}{lccc}
\toprule
 \textbf{Synthesis Model} & \textbf{Synthesis Method} & \textbf{Format Corr.} & \textbf{Summ. CE} \\
\midrule
SFT  &  One-shot & 24\% & 4.74 \\
SFT & Iterative  & 100\%* & 4.69 \\
\midrule
Joint Preference Set  & One-shot &  78\% & 4.62 \\
Joint Preference Set + SFT  & One-shot &  \textbf{96\%} & \underline{4.45} \\
Separate Preference Sets + SFT (MRDS) & One-shot &  \underline{92\%} & \textbf{4.13} \\
\bottomrule
\end{tabular}
\caption{Dialouge Synthesis with different training strategy on SAMSum 300 shots experiments.}
\label{tab:DPO abl}
\end{table*}

We evaluated the performance of different dialogue synthesis methods, including direct synthesis (One-shot), iterative dialogue synthesis (Iterative), and various configurations of DPO, as presented in Table~\ref{tab:DPO abl}. In our DPO training, we constructed different sets of preference pairs based on the definitions provided earlier. For the single joint preference set, we used dialogues from the post-processing step as the preferred data (\( \hat{d}_1 \)) and the raw dialogues without iterative synthesis as the rejected data (\( \hat{d}_2 \)), forming preference pairs:
\(\{s, \hat{d}_1, \hat{d}_2 \mid s \in \mathcal{S}, \mathcal{F}(\hat{d}_1) = 1, \mathcal{F}(\hat{d}_2) = 0, \mathcal{M}_{\text{sum}}(s \mid \hat{d}_1) > \mathcal{M}_{\text{sum}}(s \mid \hat{d}_2)\}\). 
For the separated preference sets, we included both formatting preference pairs:
\(\{s, \hat{d}_1, \hat{d}_2 \mid s \in \mathcal{S}, \mathcal{F}(\hat{d}_1) = 1, \mathcal{F}(\hat{d}_2) = 0\}\)
and summarization preference pairs:
\(\{s, \hat{d}_1, \hat{d}_2 \mid s \in \mathcal{S}, \mathcal{M}_{\text{sum}}(s \mid \hat{d}_1) > \mathcal{M}_{\text{sum}}(s \mid \hat{d}_2)\}\). 
We also incorporated supervised fine-tuning data into the DPO training, indicated as "+SFT."

From Table~\ref{tab:DPO abl}, the SFT model's one-shot synthesis had low format correctness (24\%) and a high summarization cross-entropy (CE) loss of 4.74. While Iterative synthesis ensured 100\% format correctness, the CE loss remained similar at 4.69, showing little improvement in content alignment.
Applying DPO with the joint preference set improves format correctness to 78\% and reduces the summarization CE loss to 4.62. However, training with DPO loss alone was unstable, so adding SFT loss ({Joint Preference Set + SFT}), stabilized training, achieving 96\% format correctness and a CE loss of 4.45. Yet, this approach prioritized formatting over content quality, limiting further improvement in CE loss.

By separating the preference sets and including both formatting and summarization preferences in DPO training ({Separate Preference Sets + SFT}), our MRDS approach effectively balanced format correctness (92\%) and significantly lowered the CE loss to 4.13, indicating better alignment with summaries. This demonstrates that separating preference sets allows concurrent optimization of formatting and content, leading to the best overall performance.

\subsection{Statistical Analysis of Dialogue Summarization Experiments}
To further validate the improvements introduced by the MRDS method, we conduct each experiment three times under the 100‐shot setting on both the SAMSum and DialogSum datasets. For each run, we compute the mean and sample standard deviation (using $n-1$ in the denominator) for ROUGE-1, ROUGE-2, ROUGE-L, and BERTScore. In addition, independent two-sample t-tests (with degrees of freedom $df=4$) were performed to assess the statistical significance of the differences between the baseline (summarization adapter fine-tuned with real data only) and MRDS results.

Tables~\ref{tab:stat_samsum} and~\ref{tab:stat_dialogsum} summarize the results for SAMSum and DialogSum, respectively. For instance, on SAMSum the ROUGE-1 score improved from an average of 50.90\% (std = 0.18\%) to 52.10\% (std = 0.25\%), a gain of 1.20 percentage points (t = 6.80, p $\approx$ 0.003). Similar statistically significant improvements are observed across all metrics and on both datasets.

\begin{table*}[htbp]
\centering
\caption{Statistical analysis on SAMSum 100-shot experiments (values in percentages). The baseline method adopts summarization adapter finetuned with real data only.}
\label{tab:stat_samsum}
\begin{tabular}{lccccc}
\toprule
\textbf{Metric} & \textbf{Baseline} (mean $\pm$ std) & \textbf{MRDS (Ours)} (mean $\pm$ std) & \textbf{Difference} & \textbf{t-value} & \textbf{p-value} \\
\midrule
ROUGE-1    & 50.90 $\pm$ 0.18   & 52.10 $\pm$ 0.25   & +1.20    & 6.80 & 0.003 \\
ROUGE-2    & 26.54 $\pm$ 0.04   & 27.54 $\pm$ 0.31   & +1.00    & 5.52 & 0.005 \\
ROUGE-L    & 42.62 $\pm$ 0.09   & 43.42 $\pm$ 0.35   & +0.80    & 3.87 & 0.018 \\
BERTScore  & 86.59 $\pm$ 0.04   & 86.81 $\pm$ 0.07   & +0.22    & 4.84 & 0.011 \\
\bottomrule
\end{tabular}
\end{table*}

\begin{table*}[htbp]
\centering
\caption{Statistical analysis on DialogSum 100-shot experiments (values in percentages). The baseline method adopts summarization adapter finetuned with real data only.}
\label{tab:stat_dialogsum}
\begin{tabular}{lccccc}
\toprule
\textbf{Metric} & \textbf{Baseline} (mean $\pm$ std) & \textbf{MRDS (Ours)} (mean $\pm$ std) & \textbf{Difference} & \textbf{t-value} & \textbf{p-value} \\
\midrule
ROUGE-1    & 44.04 $\pm$ 0.55   & 45.47 $\pm$ 0.15   & +1.43    & 4.35 & 0.015 \\
ROUGE-2    & 18.23 $\pm$ 0.35   & 19.26 $\pm$ 0.27   & +1.03    & 4.06 & 0.017 \\
ROUGE-L    & 35.97 $\pm$ 0.49   & 37.25 $\pm$ 0.13   & +1.28    & 4.32 & 0.016 \\
BERTScore  & 86.81 $\pm$ 0.11   & 87.19 $\pm$ 0.02   & +0.38    & 5.86 & 0.008 \\
\bottomrule
\end{tabular}
\end{table*}

The t-test results confirm that the improvements achieved by MRDS over the baseline are statistically significant (p-values < 0.05) across all metrics on both datasets.

\subsection{Comparison with State-of-the-art Methods}
In Table \ref{tab:compare_sota}, we compare the proposed MRDS with three existing state-of-the-art methods that use data augmentation and pseudo labeling to address the data scarcity problem for dialogue summarization task.
The results show that our MRDS method achieves competitive or better performance in Rouge-1, Rouge-2 and Rouge-L.
In the table, we also compare the different backbone models, the amount of few-shot paired data and the need of unlabelled data among existing methods and MRDS.

\begin{table*}[htbp]
\caption{Comparison among the proposed MRDS and three existing state-of-the-art methods (Semi-CODA \cite{chen2021simple}, SiCF \cite{he-etal-2024-semi} and COMPO \cite{ouyang2023compositional}) on SAMSum dataset. $^*$Using unlabled data.}
\label{tab:compare_sota}
\begin{tabular}{lccccc}
\toprule
Methods     & backbone       & num. of paired data & Rouge-1 & Rouge-2 & Rouge-L \\ \hline
Semi-CODA$^*$   & BART-large               & 147                 & 42.16   & 17.82   & 38.89   \\
SiCF$^*$      & DialogLM                & 147                 & 45.85   & 19.90   & 35.96   \\
COMPO       & BART-large               & 147                 & 49.78   & 24.65   & 45.41   \\
MRDS (ours) & Llama3-8B-Instruct       & 100                 & 52.1    & 27.5    & 43.4    \\ 
\bottomrule
\end{tabular}
\end{table*}

\end{document}